# An Analysis of Elephants' Movement Data in Sub-Saharan Africa Using Clustering

Gregory Glatzer[*], Prasenjit Mitra, and Johnson Kinyua

*College of Information Science and Technology, The Pennsylvania State University, United States*

[*]*Corresponding author: gvg5207@psu.edu*

## Abstract

*Understanding the movement of animals is crucial to conservation efforts. Past research often focuses on how various factors affect movement, rather than looking at locations of interest that animals often return to or habitat. We explore the use of clustering to identify locations of interest to African Elephants in several regions of Sub-Saharan Africa. Our analysis was performed using publicly available datasets for tracking African elephants at Kruger National Park (KNP), South Africa; Etosha National Park in Namibia; as well as areas in Burkina Faso and the Congo. Using the DBSCAN and KMeans clustering algorithms, we calculate clusters and their centroids to simplify elephant movement data and highlight important locations of interest. Through a comparison of feature spaces with and without temperature, we show that temperature is an important feature to explain movement clustering. Recognizing the importance of temperature, we develop a technique to add external temperature data from an API to other geospatial datasets that would otherwise not have temperature data. After addressing the hurdles of using external data with marginally different timestamps, we consider the quality of this data, and the quality of the centroids of the clusters calculated based on this external temperature data. Finally, we overlay these centroids onto satellite imagery, as well as overlaying locations of human settlements, to validate the real-life applications of the calculated centroids as a means to identify locations of interest for elephants. As expected, we confirmed that elephants tend to cluster their movement around sources of water as well as some human settlements, especially those with water holes. Identifying key locations of interest for elephants is beneficial in predicting the movement of elephants and preventing poaching. These methods may in the future be applied to other animals beyond elephants to identify locations of interests for them.*

## Introduction

Having methods to identify key locations of interest for elephants would be beneficial in predicting elephant movement and eventually help prevent poaching (Mills et al., 2018). Many methods have been used to understand elephant (and other endangered species) and poacher movement. GPS trackers (Thaker et al., 2019) and drones (Bondi et al., 2019; Hambrecht et al., 2019) are often used to collect movement data. Additionally, advanced statistical, AI, and machine learning (ML) methods have been developed in recent years (Wall J et al., 2014; Fang et al., 2016; Bondi et al., 2019; Hambrecht et al., 2019). For instance, Hambrecht et al. used computer vision techniques to identify animals and poachers in UAV drone video feeds (Hambrecht et al., 2019).

The movement data of elephants can be complex. Applying clustering algorithms to elephant movement data, however, allows the data to be simplified into fewer, more important data points. Using clustering as a form of dimensionality reduction, we can highlight centers of clusters ("centroids") as possible important locations of interest for elephants and focus on these points instead of the entire dataset. These frequented locations will be called "locations of interest", as the elephant may be coming to these spots for various reasons, such as a food source, shade, water, etc. Often, these locations of interest are driven by the presence of human settlements.

In this paper, we aim to use clustering to simplify the complex movement data and, using the centroids calculated from clustering, identify locations of interest for the elephants. These locations of interest can then be overlaid onto other data, such as human settlements, to understand where elephants may be inhabiting, and why. We will also consider how heat data contributes to the locations of interest for elephants, and thus how heat data affects the performance of the clustering algorithm. Finally, we will explore methods to approximate heat data when it was not collected in the first place. In our work, we performed clustering using two ML algorithms, DSCAN (Density-Based Spatial Clustering of Applications with Noise) and KMeans. Firstly, we overlay the centroids calculated using DBSCAN onto real-world data such as water sources and human settlements to evaluate the real-world application of the locations of interest. Finally, we will use KMeans to rank human settlements based on how many elephant locations of interest were found nearby. This gives us an end-to-end approach to calculate possible locations of interest and human settlements that elephants may be found inhabiting from raw data.

**Related work**

Using spatial data to understand the influence of elephants' surroundings on their movement has been extensively studied. Sitompul, et al., 2013 used spatial data to conclude that elephants stay near the edge of canopy coverage. Mills et al., 2018 analyzed the influence of vegetation and seasonal variables on movement. Human-Elephant Conflict (HEC) has also been addressed using spatial data. Tripathy et al., 2021 used KMeans to determine hotspots of HEC. Sitati, N.W. et al., 2003 studied how crop raiding by elephants is affected by road proximity. Granados, et. al., 2012 used spatial data to analyze the influence of human infrastructure and settlements on elephant movement, with a focus on seasonal influence and crop raiding. Beyond elephants, Lamb et al., 2020 considered the application of clustering algorithms to explain groupings of points within animal movement data. Our work, although not directly addressing HEC, contributes to the body of work, as it attempts to identify human settlements that elephants appear to be inhabiting via spatial analysis using multiple clustering algorithms.

| Dataset | Location | # Elephants | Study area coverage | # Data points | Frequency of sampling | Sensor | Published data URL |
|---|---|---|---|---|---|---|---|
| **Thaker et al., 2019** | Kruger National Park, South Africa | 14 | 31.1° E−32.0° E, 23.9° S−25.4° S | 283,737 | 30 mins | GPS collar with in-built temperature sensor | http://dx.doi.org/10.5441/001/1.403h24q5/1 |
| **Tsalyuk M et al., 2018** | Etosha National Park, Namibia | 15 | 22,270 $km^2$ | 1,370,000 | 1 to 19 mins | GPS/GSM platform collars | http://dx.doi.org/10.5441/001/1.3nj3qj45/1 |
| **Wall J et al., 2014** | Samburu National Reserve, Kenya and Gourma region, Burkina Faso | 2* | 3.1° W−0.8° E, 16.6° N−14.3° S | 23,494 | 1 hour | GPS tracking collar | http://dx.doi.org/10.5441/001/1.f321pf80/1 |
| **Blake et al., 2001** | Dzanga-Sangha and Nouabalé-Ndoki National Parks Complex of Central African Republic and Congo | 2 | 5,150 $km^2$ | 37,048 | ~22 mins | VHF (Very High Frequency) collar | https://www.movebank.org/cms/webapp?gwt_fragment=page=studies,path=study1818825 |

*Entire study recorded 9 elephants, but public dataset on Movebank only includes 2 elephants

**Table 1** Overview of data sources

### The data

Much of this analysis was performed using publicly available data from the study by Thaker et al. (Thaker et al., 2019), which took place in Kruger National Park in South Africa and consisted of 14 female African Savannah Elephants, each from a different herd. Each elephant was fitted with a GPS tracker and in-built heat sensor to measure location and external temperature at intervals of 30 minutes. They collected 283,737 GPS location points from 2007 to 2009, with "roughly equal points in the dry (n = 138,764) and wet seasons (n = 144,973)". In this paper, this data source will be referred to as the Kruger dataset.

We will also use data compiled by Tsalyuk et al. (Tsalyuk et al., 2018) in Etosha National Park in Namibia, Wall et al. (Wall et al., 2014) in the Gourma region between Mali and Burkina Faso, and Blake et al. (Blake et al., 2001) in national parks throughout central Africa and the DRC. The dataset for Wall et al. was made publicly available on Movebank, however the collection of the data is described in Wall et al. (Wall et al., 2013). These datasets were used because of their lack of a temperature feature, which was proven to be important when working with the Kruger dataset. Using these datasets with no temperature feature, we could test the performance of adding external temperature data from a weather API as a substitute for the missing important feature when performing clustering. This addition of API-sourced temperature data posed some problems, including not being able to match timestamps exactly between the datasets and the API data. This problem will be discussed in further detail in the next section.

## Methods

Real-world spatial movement data of animals often has complex patterns. Elephants, for instance, may display shuttling motion between areas of shade and water (Thaker et al., 2019), and may move long distances over time without stopping to rest in an area. The nature of this movement calls for a way to differentiate between shuttling motion and "locations of interest" in spatial data. These locations of interest can be identified as areas where lots of data points are in a dense cluster, as elephants are often returning to that location, or staying there for a prolonged period. In contrast, thin, spread-out points, which may signify the elephant moving to their next resting point, must be identified as noise. To identify locations of interest for elephants, we will use clustering. Due to the shuttling motion of elephants, it is necessary to be able to differentiate between clusters and noise. DBSCAN is an obvious fit, as it has a concept of noise in comparison to other common clustering algorithms such as KMeans, which do not. The scikit-learn implementation of DBSCAN was used, with a preprocessing step of scikit-learn's StandardScaler.

### Influence of temperature on clustering

Based on the findings of Thaker et al. (Thaker et al, 2019), who highlighted the importance of temperature to explain the shuttling motion of elephants, we considered the importance of temperature for our analysis. Accordingly, we ran DBSCAN under two feature spaces: (1) "Temp-Influenced" - temperature, latitude, and longitude, and (2) "Without Temp-influence" – latitude, longitude. The second feature space omits the temperature feature to evaluate how it influences the performance of the clustering algorithm's ability to find locations of interest. The combined results of running DBSCAN with each feature space on elephant AM306 from the Kruger dataset is illustrated in Figure 1. Large colored dots represent the centroids of the corresponding cluster, calculated using only location data. The centroids are calculated as the mean of all the points in the given cluster. Additionally, centroids calculated with temperature included in the feature space are shown as black Xs. Since the visualization does not have an axis for temperature, the Temp-influenced clusters are not visualized, but rather only the Temp-influenced centroids as black Xs, projected from the 3D space of latitude, longitude, and temperature, to the 2D space of only latitude and longitude.

Figure 1 illustrates that temperature can aid in identifying centroids in spatial elephant movement data. The "Without temp-influence" feature space was able to identify large clusters in the data but missed some more granular dense areas of points within these larger clusters. For example, the dark blue cluster in the top-left of Figure 1 has a smaller top portion that was missed using location data alone. However, by introducing temperature, the "Temp-influenced" DBSCAN was able to identify a centroid (black X) in this upper region. This infers that the density of these points may be explained by temperature. Perhaps there may be a water hole, shade, or other real-world feature that explains this location of interest.

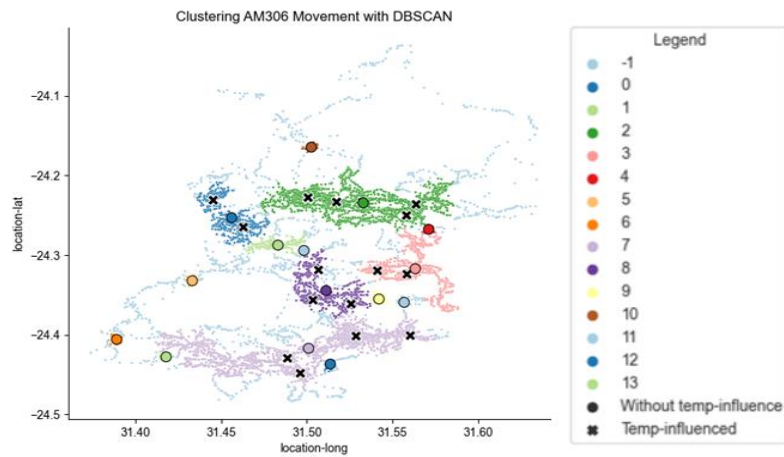

**Figure 1** Comparing feature spaces with and without temperature on elephant AM306 from the Kruger dataset. Black Xs represent centroids calculated with temperature included in the feature space. Large colored dots represent centroids of clusters calculated solely on location data. *Parameters: Without temp-influenced epsilon=0.1, minPts=35. Temp-influenced epsilon=0.2, minPts=50.*

Next, we apply DBSCAN with both feature spaces to other elephants' movement data within the Kruger dataset. The same parameters were used as before. Varying degrees of success were observed with the Temp-influenced centroids. Figure 2.1 shows a successful application, with multiple Temp-influenced centroids in the dense blue cluster that would be hard, if not impossible, to identify with location data alone. In contrast, Figure 2.2 is an instance where the temp-influenced centroids do not contribute much information.

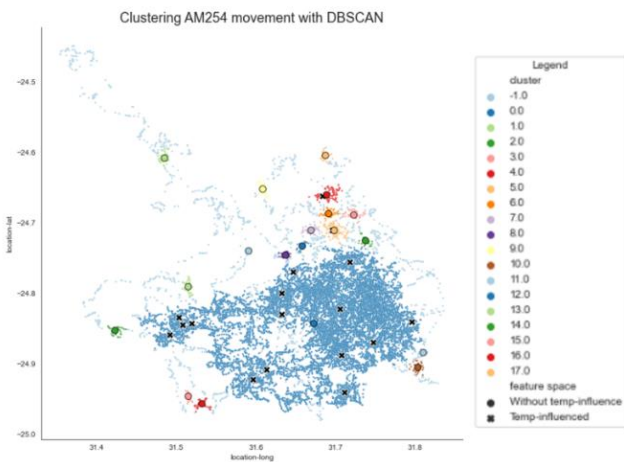 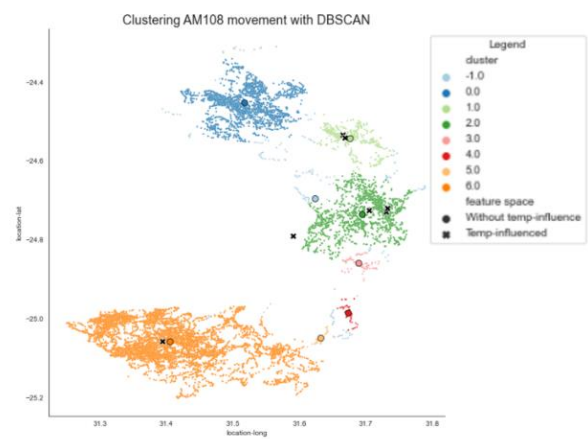

**Figure 2.1** A successful application        **Figure 2.2** A failed application

**Figure 2** More examples of Temp-influenced clustering on different elephants in the Kruger dataset.

**Adding temperature data to other datasets**

We have shown temperature data is beneficial for calculating locations of interest. However, this analysis is only possible when the temperature data is available. Other studies regarding the movement of elephants such as Tsalyuk et al. (Tsalyuk et al., 2018) and Wall et al. (Wall et al., 2014) do not provide a temperature feature. Hence, our feature space is limited to spatial data unless there is a way to approximate temperature data from other data sources. Using the meteostat python package and API, we can find weather stations whose locations are close to a study site. From there, historical data can be queried and appended to a study's data. This allows the calculation of Temp-influenced centroids that otherwise would never have been able to be calculated. This section will cover the process of collecting, validating, and implementing this data source.

There are three steps in the process of gathering and using historical weather station data for DBSCAN: (1) locate a nearby weather station to the data that temperature needs to be added, (2) match timestamps to the queried data, and (3) evaluate the ability of the appended historical temperature data to calculate temp-influenced centroids. For (1), the median of the given data's lat/long is calculated (the given data in this case is a single elephant's movement data), and the coordinate is used to query a nearby station. To achieve (2), the timeseries data from the station is normalized and interpolated. Interpolation provides higher temporal granularity in the weather station data, thus allowing more timestamps to match the given data. Both the normalization and interpolation functionality are provided by meteostat. For (3), the correlation between the historical station data and Kruger temperature data is evaluated using R-squared.

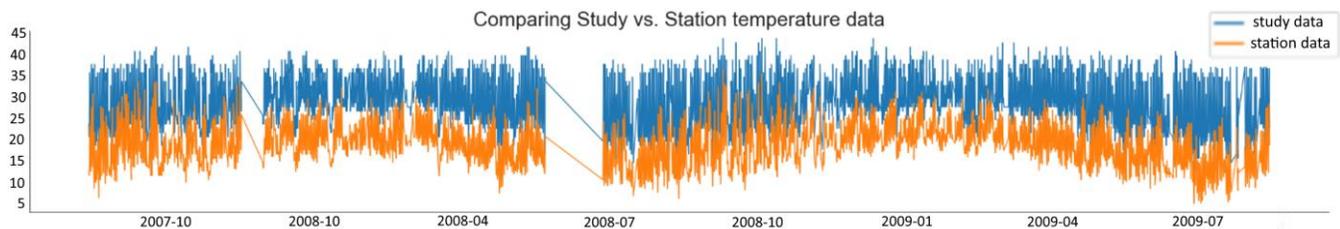

Figure 3 The difference between temperature data from Elephant AM105 in the Kruger dataset, and corresponding temperature data from the meteostat weather station API.

| R-squared (zero-centered) | 0.6871044690549571 |
|---|---|
| offset (study – station) | 9.840106696689293 |
| % of timestamps found | 61.6% |

Table 2 Statistics for Figure 3

Figure 3 and Table 2 show the results of performing steps (1)-(3) on Elephant AM105 from the Kruger dataset. The station temperature data is moderately like the study temperature data, with an R-squared of 0.6978. According to Moore et al. (Moore et al., 2013), "If R-squared value $0.5 < r < 0.7$ this value is generally considered a moderate effect size." The data was zero-centered before calculating the R-squared. The zero-centering should not matter for our case, however, as we are interested in the relationship between temperature values to calculate clusters, and not the values themselves. It is important to note the "% of timestamps found". This means that the weather station only matched temperature measurements for 61% of the original data (for this elephant). This 61% data retention is after applying both the normalization and interpolation techniques. Fuzzy timestamp matching (matching proximal timestamps within some threshold) could help mitigate this but may result in a data quality loss in exchange for more data. Fuzzy timestamp matching will be explored later. Next, we evaluate how well DBSCAN performs with station data.

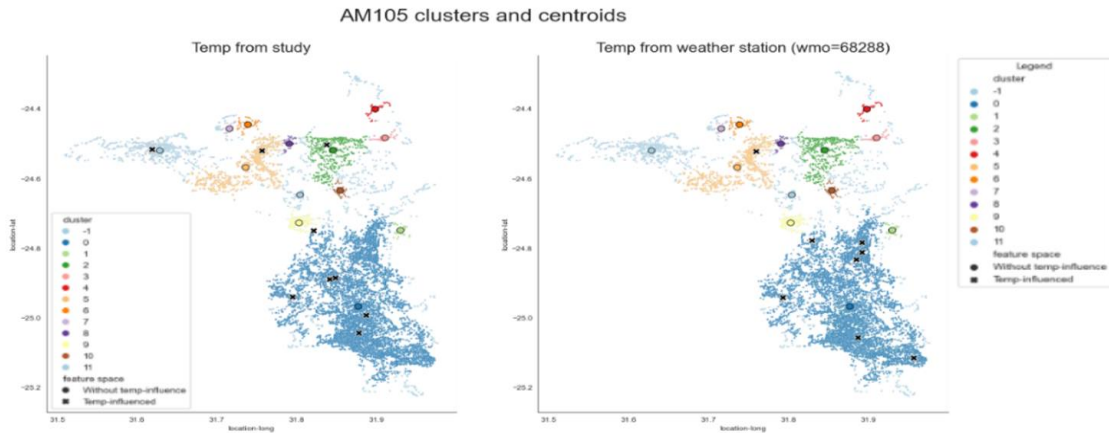

Figure 4 Applying DBSCAN to AM105 with study data and weather station data

When looking at the performance of the weather data with DBSCAN on elephant AM105 in the Kruger dataset, the Temp-influenced centroids successfully identify points of high density, i.e., the locations of interest described earlier. Interestingly, the station data identifies some same locations of interest as the original study data, but also identifies new points. It is worth noting that the same parameters as before were used for both maps. Considering the performance of the Temp-Influenced centroids with the weather data, as well as the moderate correlation between the study data and the station data, we can move forward with more confidence to apply this technique onto datasets that lack temperature data.

**Applying station data to other datasets**

The next step is to apply the meteostat techniques to datasets that do not have heat data. First, station data will be found for Etosha National Park's elephants. Due to the location and date of the study, far fewer timestamps were matched in the Etosha dataset than in the Kruger dataset. For elephant AG189 only 19.662% of the timestamps could be matched. Regardless, the limited weather station data still performed well, identifying logical Temp-influenced centroids as seen in Figure 5.

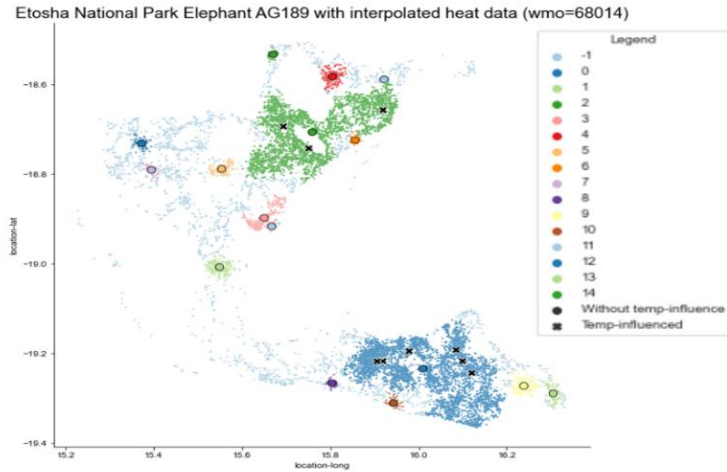

**Figure 5** Applying clustering to elephant AG189 from the Etosha dataset after appending temperature data from the meteostat weather API. *Parameters: Without temp-influenced epsilon=0.06, minPts=45. Temp-influenced epsilon=0.2, minPts=25.*

Lastly, historical station temperatures will be found for the Wall et al. (Wall et al., 2014) dataset. This dataset is interesting, as the movement patterns are vastly different compared to those seen before. As illustrated in Figure 6, these elephants move around for long distances more frequently, with smaller clusters. The Without Temp-influenced DBSCAN successfully identified logical clusters and marked the paths between them as noise (cluster -1). The heat data, however, did not provide much insight. This example illustrates that although the algorithm works on various datasets, temperature data is not always too useful.

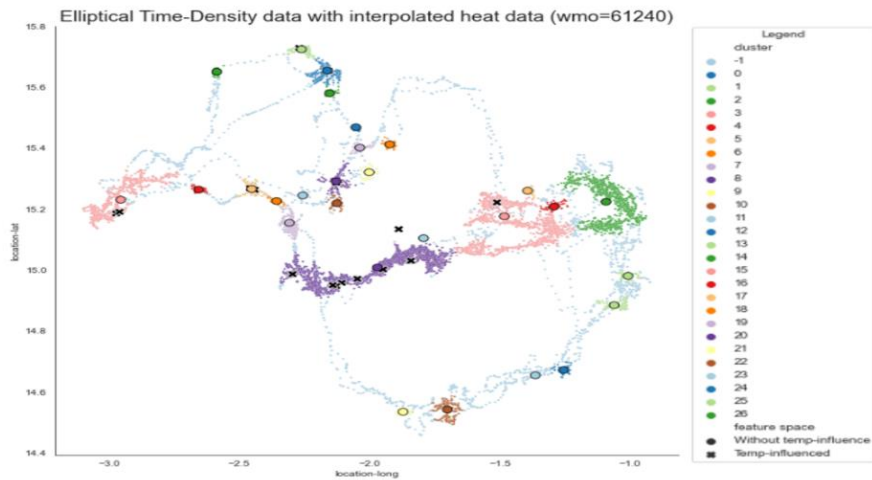

**Figure 6** Applying clustering to Habiba the elephant from Wall J et al., 2014. *Parameters: Without temp-influenced epsilon=0.06, minPts=45. Temp-influenced epsilon=0.1, minPts=35.*

**Fuzzy Timestamp matching**

A major concern when applying weather station temperature data to datasets was the small percentage of timestamps that were matched. For instance, only 19.662% of the timestamps were matched for AM189 from Etosha as we saw in Figure 5. An approach to counter this data loss is "fuzzy" timestamp matching. When joining the data from a weather station to data from a study, the data is not only joined on exact matches of timestamps, but rather a threshold is provided that two timestamps can differ within and still be considered a match in the join operation. The tolerance used was calculated by taking half of the median of the difference of timestamps in the dataset. Usually, the matched timestamps are within a threshold of 15-60 minutes.

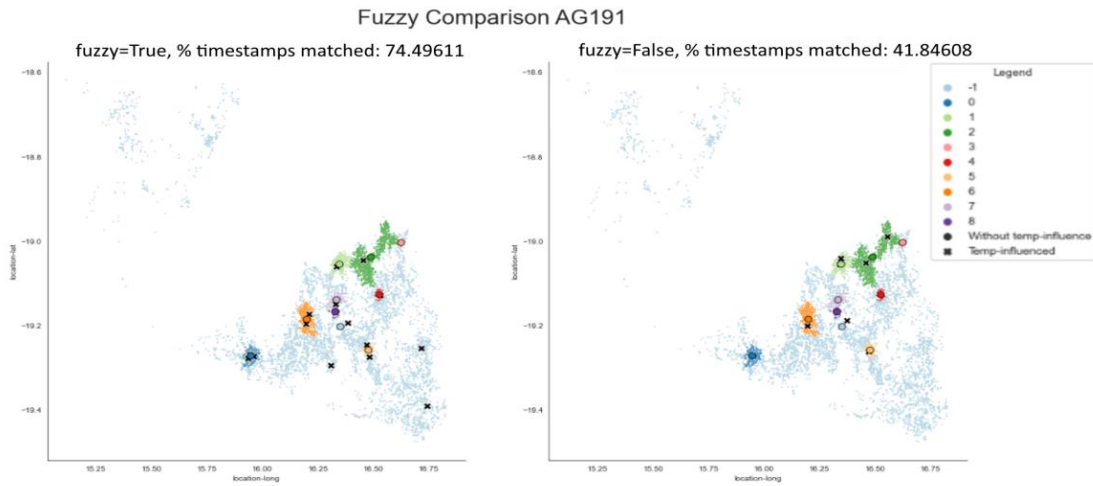

**Figure 7** Fuzzy timestamp matching results on AG191 from the Etosha dataset

Applying fuzzy matching yielded promising results, as seen in Figure 7. Using a regular join between station and study timestamps on AG191 from Etosha only matched 41.85% of the data. Applying fuzzy matching yielded 74.50% matches. These additional matched timestamps are of course not exact values but rather close times within the threshold. Using fuzzy matching, we can see many of the same Temp-influenced centroids are found, with additional centroids identified in the lower-right portion of the map. Further tests with fuzzy matching have shown similar results. There are more Temp-influenced centroids found in areas where often no Temp-influenced centroids were found beforehand. In general, it is recommended that fuzzy timestamp matching be used to find centroids.

## Discussion

Several methods have been combined to explain the movement of elephants. DBSCAN was used to calculate clusters and centroids based on location ("Without Temp-influenced"), and location plus temperature ("Temp-influenced"). Both feature spaces showed promising abilities to identify centroids in elephant movement, with each identifying some centroids that the other missed. Using a historical weather API, we added temperature data to datasets without temperature data, and showed this data is viable for our use case and can successfully be used to find centroids. Finally, we applied "fuzzy" timestamp matching to increase the number of timestamps we can find temperature for

In order to measure the value of these techniques for real world applications, additional information will need to be overlaid over the calculated centroids to put the results into context. This includes terrain, satellite imagery, human settlements, water sources, and population density. This additional information will give us an idea if the calculated centroids – Temp-Influenced and Without Temp-influenced – have any real-world meaning. We have created an application to allow the reader to explore the calculated centroids, and how they may relate to these additional features overlaid on a map. Much of the discussion comes from analyzing the centroids through the application, which can be found at https://share.streamlit.io/g1776/elephantcentroids/main/app.py.

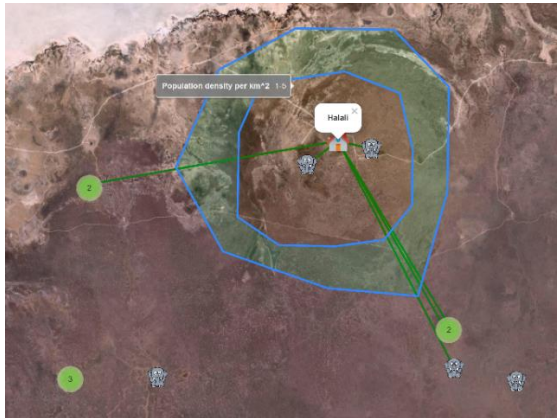
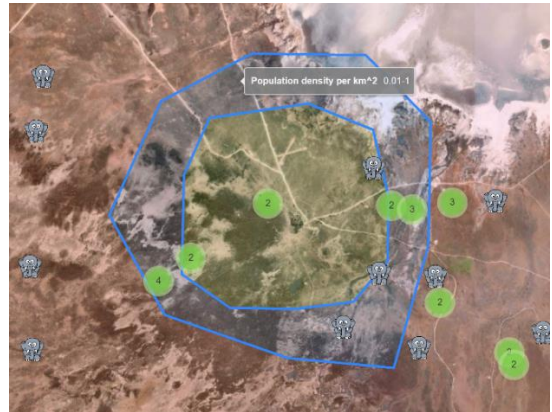

**Figure 8.1** Halali campsite  **Figure 8.2** Okaukuejo campsite
**Figure 8** Elephant centroids around human settlement in Etosha National Park

Figure 8 shows elephant centroids near human settlements in Etosha National Park at the Halali and Okaukuejo campsites. Centroids are represented as elephant icons (from http://clipart-library.com/), with green markers representing groups of centroids. These centroids consist of both Temp-Influenced and without Temp-influenced variants using fuzzy matching. The ring around the settlement is the extent at which humans populate the area. Centroids appear around and in the ring, inferring that elephants may have interest in the settlement and the surrounding area. According to Etosha National Park's website, both the Halali and Okaukuejo campsites have a watering hole, which very well could be the reason these elephants have inhabited these areas (Halali Camp; Okaukuejo Camp). We have found also found centroids to appear near bodies of water. In Figure 9, elephant centroids from the Kruger dataset are seen along a river, with a camp seen slightly down the river. Perhaps the elephants care more about the presence of the river than caring about the settlement, as there are not many elephant centroids focused around the settlement, but rather along the river.

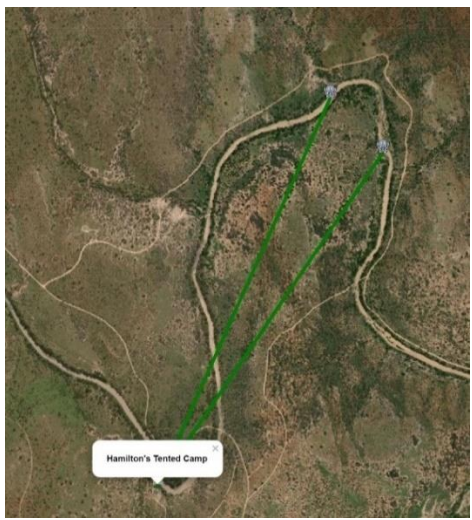
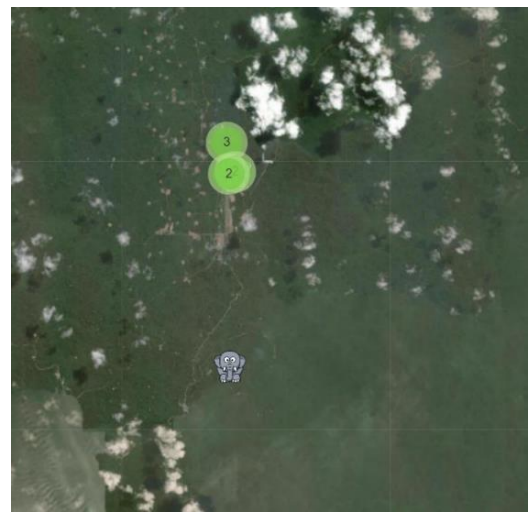

**Figure 9** Elephants along a river (Kruger dataset)  **Figure 10** Elephants near Rabi Kounga Oil Field (Blake et al., 2001)

In Figure 10, a cluster of elephant centroids is found around a settlement. This settlement is the Rabi Kounga Oil Field. Not much information could be found regarding the nature of this settlement, or why elephants may have interest in it. Regardless, there is a dense cluster of many centroids around this settlement with no other centroids for many miles, which infers that elephants have expressed interest in this settlement. Overall, putting the centroids into the context of their surroundings helped us better understand how they can be useful in a real application. We saw that some centroids are at or around settlements with watering holes, while others tend to be near rivers. In analyzing the centroids, we failed to notice any pattern that would infer that the Temp-influenced centroids were any different than the without Temp-influenced centroids. We were expecting to see Temp-influenced centroids close to water sources and clusters of trees (shade) more often than Without Temp-influenced centroids, however this was not the case.

### Using KMeans to automate finding important human settlements

Looking at the combined centroids plus human settlements gives us the ability to find human settlements that elephants are potentially inhabiting at times. It is easy for a human to look at the map and identify which settlements have lots of elephant centroids and should be investigated further. We can automate the selection process by applying KMeans and DBSCAN algorithms to the problem We then ranked the settlements based on how many elephant centroids are in each respective cluster. The result of this process ran on the Etosha dataset is shown in Table 3. This gives a good starting point of where to direct efforts to find elephants.

### Conclusion

Performing spatial clustering on elephant data may be helpful in identifying locations of interest for elephants. In this paper, DBSCAN was applied onto elephant datasets from four studies as explained earlier. We explored DBSCAN's ability to identify clusters in these datasets, and how different parameters in this algorithm can affect the results.

| Geometry | Name | Type | # centroids in settlement cluster |
|---|---|---|---|
| POINT (16.4710969 -19.0356338) | Halali | village | 23 |
| POINT (15.295068 -17.6750468) | Ogongo | village | 10 |
| POINT (14.5680119 -17.3940925) | Omahenene | village | 9 |
| POINT (16.0210914 -17.9842151) | Olukonda | hamlet | 9 |
| POINT (14.3015546 -19.8967052) | - | hamlet | 1 |
| POINT (15.9146617 -17.2211624) | Okawe | village | 1 |
| POINT (14.4043373 -18.7978984) | - | village | 1 |
| POINT (14.1598649 -18.6311834) | Otjitundua | village | 1 |
| POINT (14.960421 -20.37384) | Khorixas | town | 1 |
| POINT (17.4722164 -17.6549121) | Omupini | hamlet | 1 |

**Table 3** Ranking human settlements in Etosha based on number of DBSCAN-calculated centroids around them.

In addition to a basic application of DBSCAN onto a spatial dataset, we evaluated the ability of a temperature feature to improve the performance of clustering on elephant movement data. We attempted to match temperature data to datasets with latitude, longitude, and timestamp for each data point. We were able to match temperature readings gathered from historical weather station data via timestamps in a dataset without a temperature feature. This allowed us to utilize temperature as another feature to help explain the clustering of elephant movement. A "fuzzy" timestamp matching technique was used to increase the number of timestamps that temperature data could be found for. The resulting centroids were then analyzed in the context of their surroundings to understand if the centroids have a real-world application, and if they represent possible locations of interest for elephants.

Overall, we created an end-to-end approach from loading data to identifying potential human settlements to be further investigated. Further work may include applying this technique to animal movement other than the African Elephant (Loxodonta Africana). Datasets such as Abrahms et al. (Abrahms et al., 2017) may allow this analysis. Additionally, other methods to prevent the loss of data seen above should be further pursued. This could be combining data from multiple weather stations or finding other sources of temperature readings. The code used to generate the visualizations in this research paper, as well as the implementation code for the DBSCAN, weather station, and fuzzy timestamp matching techniques can be found at https://github.com/g1776/ElephantsDBSCANResearch.